\documentclass[letterpaper, 10 pt, conference]{ieeeconf}  

\IEEEoverridecommandlockouts                              
\usepackage[utf8]{inputenc}
\usepackage[T1]{fontenc}
\usepackage{graphicx}
\usepackage[namelimits]{amsmath} 
\usepackage{amssymb}             
\usepackage{amsfonts}            
\usepackage{mathrsfs}            
\usepackage{subfigure} 
\usepackage{booktabs}
\usepackage{cite}
\usepackage{url}
\usepackage{dsfont}
\usepackage{algorithm}
\usepackage{algpseudocode}
\usepackage{algorithmicx}
\usepackage{amsmath}
\usepackage[switch]{lineno}

\usepackage[colorinlistoftodos,bordercolor=blue,backgroundcolor=blue!20,linecolor=blue,textsize=scriptsize]{todonotes}

\title{\LARGE \bf
Improved Deep Reinforcement Learning with Expert Demonstrations for Urban Autonomous Driving}

\author{Haochen Liu, Zhiyu Huang, Jingda Wu, and Chen Lv$^{*}$,~\IEEEmembership{Senior Member, IEEE} 
\thanks{H. Liu, Z. Huang, J. Wu and C. Lv are with the School of Mechanical and Aerospace Engineering, Nanyang Technological University, 639798, Singapore. (E-mails: {\tt\small haochen002@e.ntu.edu.sg}, {\tt\small zhiyu001@e.ntu.edu.sg}, {\tt\small lyuchen@ntu.edu.sg})}
\thanks{This work was supported by the SUG-NAP Grant (No. M4082268.050) of Nanyang Technological University, Singapore.}
\thanks{$^{*}$Corresponding author: C. Lv}
}

\begin{document}
\maketitle
\thispagestyle{empty}
\pagestyle{empty}

\begin{abstract}
Learning-based approaches, such as reinforcement learning (RL) and imitation learning (IL), have indicated superiority over rule-based approaches in complex urban autonomous driving environments, showing great potential to make intelligent decisions. However, current RL and IL approaches still have their own drawbacks, such as low data efficiency for RL and poor generalization capability for IL. In light of this, this paper proposes a novel learning-based method that combines deep reinforcement learning and imitation learning from expert demonstrations, which is applied to longitudinal vehicle motion control in autonomous driving scenarios. Our proposed method employs the soft actor-critic structure and modifies the learning process of the policy network to incorporate both the goals of maximizing reward and imitating the expert. Moreover, an adaptive prioritized experience replay is designed to sample experience from both the agent's self-exploration and expert demonstration, in order to improve sample efficiency. The proposed method is validated in a simulated urban roundabout scenario and compared with various prevailing RL and IL baseline approaches. The results manifest that the proposed method has a faster training speed, as well as better performance in navigating safely and time-efficiently. 
\end{abstract}


\section{INTRODUCTION}
\label{sec1}
Autonomous driving in urban scenarios remains a major challenge today, primarily due to the complicated driving conditions, including great variance of traffic density and agent interactions, as well as the requirement of balance between efficiency (speed), comfort (smoothness), and safety \cite{new3, huang2021driving}. Current motion control strategies focus on rule- or model-based methods. These methods excel in interpretability but are with several inherent drawbacks. First of all, the rules or models are designed manually with potentially inaccurate assumptions, thus making it hard to scale to complicated real-world environments. Moreover, the rules themselves are hard to define and maintain for continual improvement.


On the other hand, with diverse and large-scale driving data, the learning-based methods that can handle large state and action spaces and complicated situations in urban driving scenarios are becoming viable. Essentially, there are two paradigms to learn the motion control strategies, namely imitation learning (IL) and reinforcement learning (RL). For IL, suppose that the trajectories from an expert demonstration are close to optimal, then IL can effectively learn to approximate the expert driving policy by reproducing expert actions given states, which could guarantee a lower bound of performance. For RL, the agent interacts with the environment and aims to optimize long-term rewards and make better decisions using its collected experiences. However, both these two paradigms have their own disadvantages. For IL, its ability is limited since its performance can at best amount to expert demonstrations. Besides, IL can easily encounter the distributional shift problem \cite{14} because it only relies on static datasets. For RL, it suffers from poor data efficiency and in complex driving situations, the reward is sparse and hard to specify. To mitigate the problems in both RL and IL and facilitate the development of the learning-based motion control strategy, a method that integrates reinforcement learning and imitation learning should be considered. Combining the merits from RL with ones from IL, reinforcement learning from demonstration (RLfD) \cite{15} is expected to not only accelerate the initial learning process of RL with the help of expert data, but also gain the potential of surpassing the performance of experts.


In this paper, we propose a novel framework combining reinforcement learning and expert demonstration to learn a motion control strategy for urban scenarios. The experiments are carried out in a roundabout scenario in the high-fidelity CARLA driving simulator. To reduce the complexity, we decompose the motion control of the vehicle into lateral control governed by a pure-pursuit controller for path tracking and longitudinal control enabled by the proposed learning framework. We first collect human driving data as the expert demonstration dataset and then utilize the designed RLfD method, which samples the experiences from both the expert demonstrations and the agent's self-exploration, to learn a motion control strategy. The results reveal that the proposed framework can effectively accelerate the learning process and accomplish better performance. The contributions of this paper are listed as follows.

\begin{enumerate}
\item A soft-actor-critic-based reinforcement learning from demonstration method is proposed, in which the learning process of the policy network is modified to accordingly combine maximizing the Q-function and imitating the expert demonstration.

\item A dynamic experience replay is proposed to adaptively adjust the sampling ratio between the agent's self-exploration and the expert's demonstration in the learning process.

\item A comprehensive test is carried out in a simulated urban driving scenario, where various RL and IL baselines are compared against our proposed method.
\end{enumerate}

\section{RELATED WORK}
\subsection{Deep reinforcement learning}
Significant progress in deep reinforcement learning (DRL) has been made in recent years, expanding its applications in a wide variety of domains, especially in the field of robotics. Previous works have shown that the model-free DRL-based approach is promising for applications in  autonomous learning motion control strategies, and thus much effort has been put on DRL-based methods. Zhang \textit{et al.} implemented a vehicle speed control strategy using the double deep Q-network (DDQN) that utilizes visual representation as system input \cite{18}. Chen \textit{et al.} investigated four model-free DRL algorithms for motion control in a roundabout scenario, namely DDQN, deep deterministic policy gradient (DDPG), twin delayed DDPG (TD3), and soft actor-critic (SAC) \cite{11}, and the results demonstrated that the SAC algorithm outperforms the others. DRL has also been applied in dedicated motion control modules on the vehicle, which delivers favorable and robust performance. Chae \textit{et al.} adopted DQN with a carefully-designed reward function for an adaptive braking system, which could effectively avoid collisions \cite{23}. Ure \textit{et al.} successfully introduced DRL to tune the parameters in MPC \cite{24} controllers, in order to achieve better and more stable performance in path tracking. Some of these DRL algorithms will be used as the baselines to show the effectiveness of our proposed method.

\subsection{Imitation learning}
In addition to the DRL method, a renewed interest in imitation learning (IL) for autonomous driving has been raised. The behavior cloning method has become the prevailing method in end-to-end autonomous driving thanks to the significant improvement brought by deep neural networks. Xu. \textit{et al.} proposed a combination of the fully convolutional network (FCN) and long-short term memory (LSTM) network for learning driving policy \cite{27}. Codevilla \textit{et al.} designed a conditional imitation learning framework which incorporates navigational command inputs \cite{28}. Huang \textit{et al.} presented a multimodal sensor fusion-based end-to-end driving system with imitation learning and scene understanding \cite{huang2020multi}. Besides, some more sophisticated imitation learning methods have been proposed recently, including generative adversarial imitation learning (GAIL) \cite{30}, and Soft-Q imitation learning (SQIL) \cite{31}, in order to enable the agent to learn from demonstrations in a more effective way. These methods are used as the imitation learning baselines in the experiment. 


\subsection{Reinforcement learning from demonstrations}
Both RL and IL have inherent downsides, as stated in Section \ref{sec1}, which gives rise to the concept of combining RL with IL for a more efficient learning process, i.e., reinforcement learning from demonstrations (RLfD). For example, Liu \textit{et al.} \cite{33} utilized DDPG with expert demonstration data in track following in The Open Racing Car Simulator (TORCS). Liang \textit{et al.} brought forward controllable imitative reinforcement learning (CIRL) built upon DDPG and imitation learning for urban navigation \cite{34}. Deep imitative model \cite{35} proposed by Rhinehart \textit{et al.} combined R2P2 and imitation learning to improve goal-directed planning. Our method is closely related to deep Q-learning from demonstrations (DQfD) \cite{15} and DDPG from demonstrations (DDPGfD) \cite{16}, which incorporate small sets of expert demonstration data into experience replay and thus show a massively accelerated training process and better performance. However, our proposed method is based on SAC, which employs a maximum entropy and probabilistic setting rather than a deterministic manner (DDPG-based method). Besides, an adaptive experience sampling method is proposed to dynamically adjust the learning objectives.

\begin{figure*}[ht]
    \centering
    \includegraphics[width=0.8\linewidth]{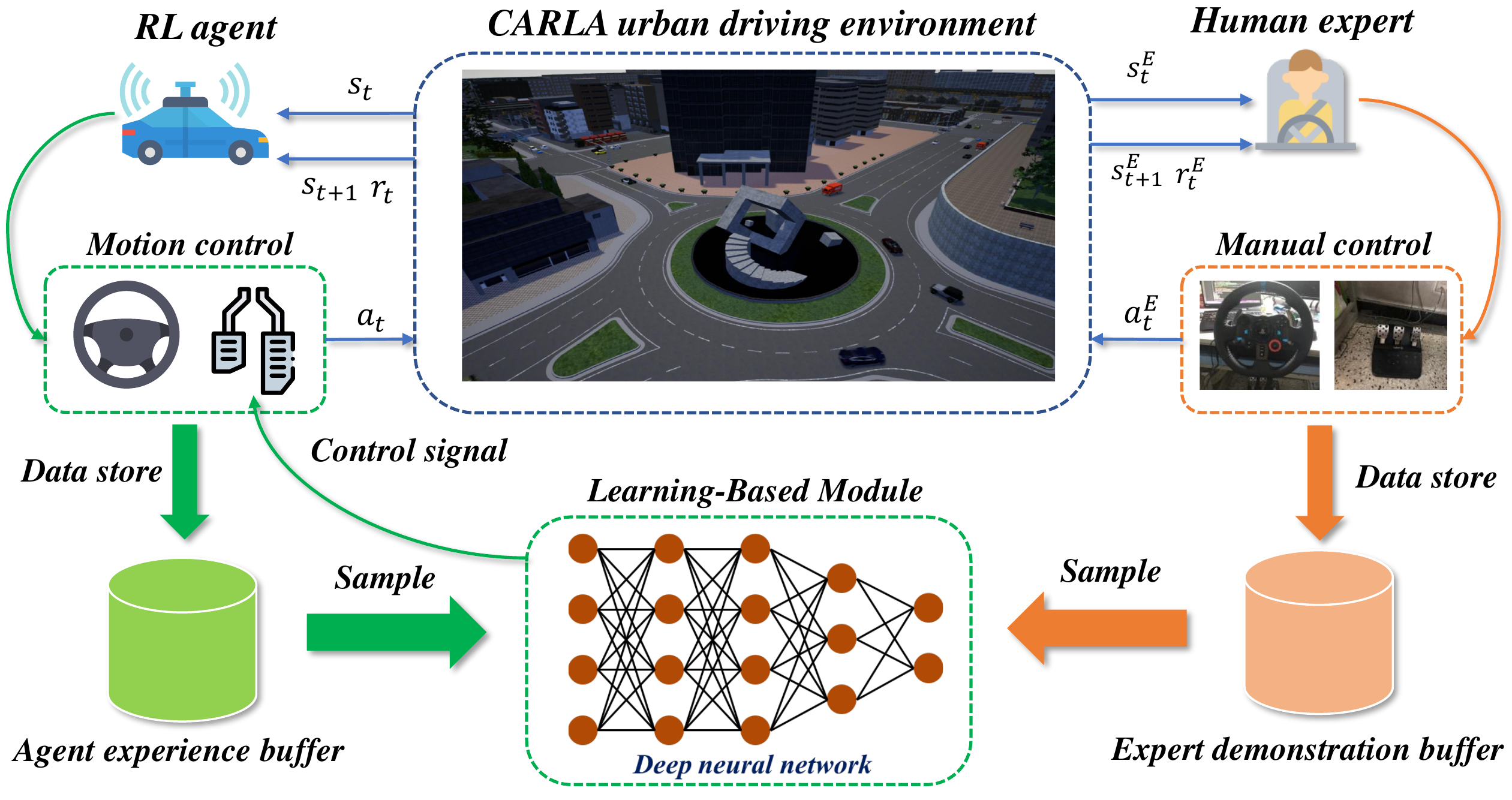}
    \caption{An overview of our motion control system. The learnable module (policy network) receives bird-eye observations and generates throttle command. The lateral motion controller generates the steering command adaptively according to the throttle, predefined route, and vehicle states.}
    \label{fig:fig.1}
\end{figure*}

\section{Method}
\subsection{Learning framework}
As shown in Fig. \ref{fig:fig.1}, the learning framework is composed of three main parts. The simulated driving environment receives the actions of the agent or human expert and then emits the states of the environment. A human expert drives in the same environment using the steering wheel and pedals and their actions are collected in the demonstration dataset. For the RL driving agent, its motion control strategy is explicitly decomposed into longitudinal (acceleration) control with neural network (learnable) and lateral (steering) control with a pure-pursuit controller (non-learnable) that tracks the predefined route. 

The learning process follows the discrete-time Markov decision process (MDP). At every timestep $t$, the agent receives the state $s_{t}$ of the environment and executes an action $a_{t}$ according to its policy $\pi(a_{t}|s_t)$, and the environment returns a reward $r_t$ and transitions to the next state $s_{t+1}$. The goal of RL is to optimize the policy to maximize the long-term expected returns: $ \max\mathbb{E}_{\pi} \left[ \sum_{t} \gamma^{t} r\left(s_{t},a_{t}\right) \right]$. On the other hand, the goal is IL is to imitate the expert demonstrations, which can be represented as minimizing the discrepancy (L2 norm) between the policy's action and the expert's action: $\min\mathbb{E}_{\pi} \left\| \pi \left(a_t|s_t^{E}\right) - a_t^{E}\right\|_{2}$. The core idea of our proposed approach is to add the IL target when using RL to train the motion control policy, in order to accelerate the training process and achieve better performance.

We maintain two experience replay buffers to store the agent's self-exploration experiences $\mathcal{D}^{S}$ and expert demonstrations $\mathcal{D}^{E}$ separately. Both $\mathcal{D}^{S}$ and $\mathcal{D}^{E}$ are stored in the format of state-action pairs along with the reward and state transition: $\left\{ \left(s_{t}, a_{t}, r_t, s_{t+1} \right)\right\}$. In the training process, the policy $\pi_{\phi}$ parameterized by $\phi$, samples a batch of mixed experiences from those two buffers with an adaptive ratio at each gradient update step. An off-policy DRL (SAC) learner integrated with imitation learning is introduced to carry out the training and the details of the learning process are given in the following subsections.

\subsection{Soft actor-critic with imitation learning}
We implement the SAC algorithm with automatic entropy adjustment \cite{haarnoja2018soft}, augmented by the proposed experience replay mechanism and imitation learning objective, to train the policy network. SAC optimizes a stochastic policy in an off-policy manner and combines the actor-critic framework with the maximum entropy principle, which helps mitigate the issues on exploration-exploitation. The algorithm concurrently learns a policy network $\pi_\phi$, a twin Q-function networks $Q_{\theta_1}, Q_{\theta_2}$ for reduction in Q-value variances, and a value function network $V_{\psi}$. The twin Q-function networks are updated according to the following loss function:
\begin{equation}
\label{eq3}
\mathcal{L}_{Q}(\theta_i) = \underset{(s_{t}, a_{t}, r_t, s_{t+1}) \sim \mathcal{D}}{\mathbb{E}} \left[ \left( Q_{\theta_i}\left(s_{t}, a_{t}\right) - y_Q \right)^{2} \right],
\end{equation} 
where $i=1,2$ and $\mathcal{D}$ is composed of the experience collected by both the agent's self-exploration and the expert's demonstration $\mathcal{D} = \mathcal{D}^E \cup \mathcal{D}^S$. The target  $y_Q$  for Q-function update is given by:
\begin{equation}
\label{eq5}
y_Q = r_t+ \gamma V_{\text{target}} \left( s_{t+1} \right),
\end{equation}
where $V_{\text{target}}$ is the target value function network, which is obtained by Polyak averaging the value network $V_{\psi}$ parameters at each gradient step.

The value function network $V_{\psi}$ gets update through the following loss function:
\begin{equation}
\label{eq6}
\mathcal{L}(\psi) = \underset{s_{t} \sim \mathcal{D}}{\mathbb{E}} \left[ \left( V_{\psi}\left(s_{t}\right) - y_V \right)^{2} \right],
\end{equation}
and the target for value function is given by:
\begin{equation}
\label{eq7}
y_V = \min_{i=1,2} Q_{\theta_i} (s_{t}, \tilde{a}_{t}) - \alpha \log \pi_\phi(\tilde{a}_{t}|s_{t}),
\end{equation}
where $\alpha$ is a non-negative temperature parameter that controls the trade-off of the entropy term. The parameter is automatically tuned over the course of training according to \cite{haarnoja2018soft}. The actions are obtained from the current policy $\tilde{a}_{t} \sim \pi_\phi(\cdot|s_{t})$, where the states are sampled from the replay buffer $s_{t} \sim \mathcal{D}$.

The agent explores the environment according to its stochastic policy, i.e., ${a}_{t} \sim \pi_\phi(\cdot|s_{t})$, and the exploitation and exploration trade-off is controlled by the entropy of the policy, e.g., increasing entropy results in more exploration. In practice, we use the Gaussian policy and thus the policy network outputs two values that represent the mean and standard deviation of a Gaussian distribution, i.e., ${a}_{t} \sim \mathcal{N} \left( \mu_\phi(s_t), \sigma_\phi(s_t) \right)$. To make the stochastic policy differentiable, we use the reparameterization trick, in which a sample of actions from the stochastic policy is drawn by computing the following deterministic function:
\begin{equation}
\tilde{a}_t = \tanh \left( \mu_\phi(s_t) + \sigma_\phi(s_t) \odot \xi \right), \ \xi \sim \mathcal{N}(0,I),
\end{equation}
where $\xi$ is independent Gaussian noise and $\tanh$ is used to ensure that actions are bounded to a finite range.

We can utilize both imitation learning and reinforcement learning to optimize the policy network, leveraging the experiences from the expert demonstration and the agent's exploration, respectively. For reinforcement learning, the policy network should be updated to maximize the expected future return plus expected future entropy. The loss function for the policy network $\pi_\phi$ when learning from the agent's experience should be:
\begin{equation}
\label{eq8}
\begin{aligned}
\mathcal{L}(\phi)_{RL} = \underset{s_t \sim \mathcal{D}^S}{\mathbb{E}} \ [ \alpha \log \pi_{\phi} \left( \tilde{a}_t | s_t \right)  \\ 
- \min_{i=1,2} Q_{\theta_{i}} \left(s_t, \tilde{a}_t \right) ].
\end{aligned}
\end{equation}

For learning from expert demonstrations or imitation learning, the loss function for updating the policy network $\pi_\phi$ becomes:
\begin{equation}
\label{eq9}
\mathcal{L}(\phi)_{IL} = \underset{(s_t, a_t) \sim \mathcal{D}^E}{\mathbb{E}} \left[ \left( \tilde{a}_\phi(s_t)-a_{t} \right)^{2} \right],
\end{equation}
where $\mathcal{D}^E$ is the buffer that contains expert demonstrations.

In the imitation loss function Eq. (\ref{eq9}), the agent's action $\tilde{a}_{\phi} \left( s_{t} \right)$ is the mean value of the stochastic policy $\tanh \left( \mu_\phi(s_t) \right)$, instead of a sample from the action distribution. In addition, we add a Q-value regularization so as to combat the overfitting issue and boost the learning speed, which means the imitation loss is only referred to if the following condition is satisfied:
\begin{equation}
\label{eq10}
Q_{\theta_j}\left( s_{t}^E, a_{t}^E \right) \ge \min_{i=1,2} Q_{\theta_i} \left(s_{t}^E, \tilde{a}_t \right), \tilde{a}_t\sim\pi_\phi(\cdot|s^E_{t}),
\end{equation}
where $j=1, 2$, meaning any of the two Q-function networks can trigger the condition.

Eq. (\ref{eq10}) illustrates that the policy network will cease imitation loss update when the agent's performance (Q-value) outweighs the expert. Empirically speaking, adding this constraint can effectively adjust the imitation learning process and filter the suboptimal demonstrations, and avoid overfitting the policy to expert demonstrations.

Putting all together, the loss function for the policy network $\pi_\phi$ in SAC with imitation learning is:
\begin{equation}
\label{eq11}
{\cal L} \left( \phi \right) = \left\{ 
\begin{array}{l}
\mathcal{L}(\phi)_{RL}, \ {\text{if} \ s_t \sim \mathcal{D}^S },\\
\mathcal{L}(\phi)_{IL}, \ {\text{if} \ (s_t, a_t) \sim \mathcal{D}^E \ \text{and Eq. (\ref{eq10})}}.
\end{array} \right.
\end{equation}

where $\mathcal{L}(\phi)_{IL}$ would be masked for expert data pairs unsatisfied by Eq. (\ref{eq10}). To balance the sources of experience the agent learns from (i.e., the ratio of the agent's experience and expert's demonstration), we design a mechanism of experience replay that can adaptively sample experiences from the two sources, which is explained below.


\subsection{Adaptive prioritized experience replay}
Following the prioritized experience replay (PER) mechanism \cite{37}, each transition tuple in the two replay buffers will be assigned a priority, such that more important transition tuples with greater approximation errors can more likely be sampled. This way, the sampling process becomes more efficient and goal-directed. The devised priority $p_{i}$ for transition tuple $i$ in the agent replay buffer is:
\begin{equation}
\label{eq12}
p_{i}^{RL} = \mathcal{L}_\pi(\phi)_{RL} + \bar{\mathcal{L}_Q({\theta})}+ \epsilon,
\end{equation}
where $\epsilon$ is a small positive constant to ensure that all transitions are sampled with some non-zero probability. $\bar{\mathcal{L}_Q({\theta})}$ indicates the mean value of the twin-Q losses. Likewise, the priority $p_{i}$ for transition tuple $i$ in the expert replay buffer is
\begin{equation}
\label{eq13}
p_{i}^{IL} = \mathcal{L}_\pi(\phi)_{IL} + \bar{\mathcal{L}_Q({\theta})} + \epsilon.
\end{equation}

The probability for a transition tuple being sampled is $ P(i) = \frac {p_{i}^{\omega}} {\sum_{k} p_{k}^{\omega}}$, in which $\omega$ is a hyper-parameter that determines the level of prioritization. To correct the bias introduced by not uniformly sampling during backpropagation, an importance-sampling weight is assigned to the loss regarding the transition tuple $w(i) = (\frac{1}{N P(i)})^{\beta}$, where $N$ is the number of experience tuples in the buffer and $\beta$ is another hyper-parameter that controls how much prioritization to apply. 

We sample the experiences separately for expert and agent buffers while updating the priorities of them, and the ratio of the samples from the two sources is dynamically adjusted using the following equation:
\begin{equation}
\label{eq14}
\mathcal{B} \leftarrow \left( {\rho \mathcal{B}} \sim \mathcal{D}^{S} \right) \cup \left( {(1-\rho) \mathcal{B}} \sim \mathcal{D}^{E} \right),
\end{equation}
where $\mathcal{B}$ is a mini-batch, and $\rho \in [0, 1]$ is the sampling ratio clipped by 0 to 1, which is updated after an episode is done according to:
\begin{equation}
\label{eq15}
\rho \leftarrow  \rho + \frac{1}{N_{\mathcal{B}}} \mathds{1} \left( \sum_{t} r_{t}^{A} \ge \overline{r}^{E} \right)
\end{equation}
where $N_{\mathcal{B}}$ is the size of the mini-batch, $\sum_{t} r_{t}^{A}$ denotes the episodic reward of the agent, and $\overline{r}^{E}$ represents the average episodic reward of the expert demonstration. It indicates that the sampling ratio for the agent buffer will gradually increase if the episodic reward for the agent is greater than the average performance of the expert. For a more specific illustration of the learning framework, the pseudo-code implementation is given in Algorithm \ref{Ag}.

\begin{algorithm}[htb]
	\vspace{-\topsep}
	\caption{Soft Actor-Critic with Imitation Learning}
	\begin{algorithmic}[1] 
	\Require Expert demonstration buffer $\mathcal{D}^E$, initial sampling ratio $\rho$, initial entropy parameter $\alpha$, Polyak averaging weight $\lambda$.
	\State Initialize policy network $\phi$, value network $\psi$, and Q networks $\theta_i,i=1,2$
	\State Initialize target value network $V_{target} \leftarrow V_\psi$
	\State Initialize empty agent replay buffer $\mathcal{D}^S$
	\Repeat
		\State Observe state $s_t$ and sample action ${a}_{t} \sim \pi_\phi(\cdot|s_{t})$
		\State Execute the action ${a}_{t}$ in the environment
		\State Observe the next state $s_{t+1}$ and reward $r_t$
		\State Store transition $(s_t,a_t,r_t,s_{t+1})$ in agent buffer $\mathcal{D}^S$
		\If {time to update}
		    \State Sample a mini-batch $\mathcal{B}$ using Eq.(\ref{eq14})
    		\State Update Q networks $\theta_i$ using Eq.(\ref{eq3}) and Eq.(\ref{eq5})
    		\State Update value network $\psi$ using Eq.(\ref{eq6}) and Eq.(\ref{eq7})
    		\State Update $V_{target} \leftarrow\ \lambda \psi +(1-\lambda)V_{target}$
    		\State Update policy network $\phi$ using Eq.(\ref{eq11}) 
    		\State Update entropy parameter $\alpha$ 
    		\State Update priorities of transition tuples $p_i^{RL}$ and $p_i^{IL}$ 
    		\State using Eq.(\ref{eq12}) and Eq.(\ref{eq13}) respectively
    	\EndIf
    	\If {$s_{t+1}$ is terminal}
    	\State Reset environment state
    	\State Update sampling ratio $\rho$ using Eq.(\ref{eq15})
    	\EndIf
    \Until{Convergence}
	\end{algorithmic}
	\vspace{-\topsep}
	\label{Ag}
\end{algorithm}

\section{Experiments}
\subsection{Experimental Setup}
\begin{figure}[htp]
    \centering
    \subfigure[]{
    \begin{minipage}[t]{0.544\linewidth}
    \centering
    \includegraphics[width=\linewidth]{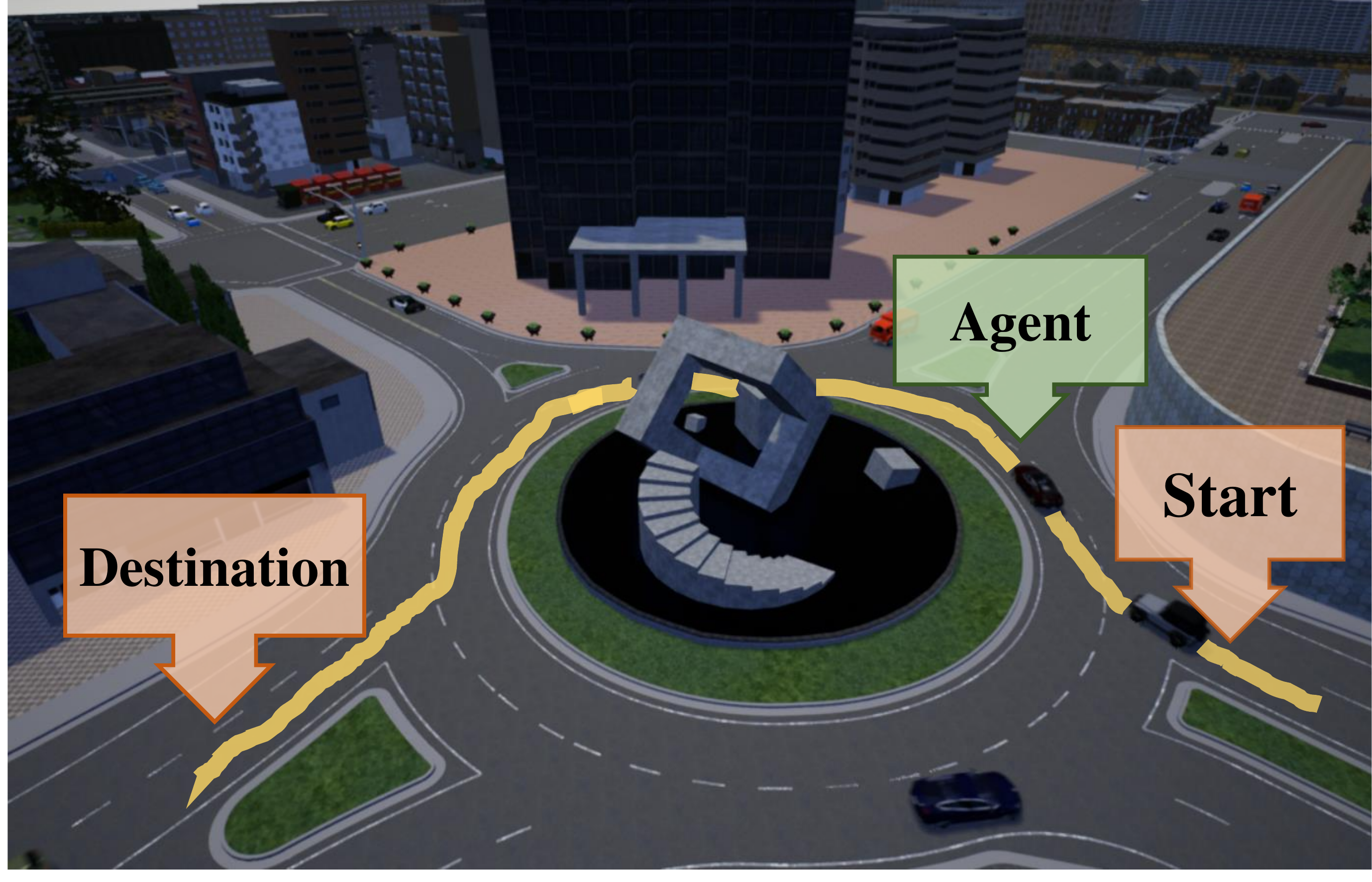}
    \end{minipage}%
    }%
    \subfigure[]{
    \begin{minipage}[t]{0.39\linewidth}
    \centering
    \includegraphics[width=\linewidth]{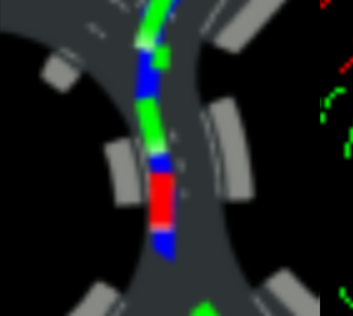}
    \end{minipage}%
    }%
    \caption{Overview of the urban driving task and processed scene representation: (a) overview of the roundabout; (b) an example of bird-view image input.}
    \label{fig:fig.3}
\end{figure}

We use the CARLA simulator \cite{38} as the experimental platform, where the autonomous vehicle is tasked to run safely and time-efficiently through a roundabout consisting of multiple intersections in the urban area, shown in Fig. \ref{fig:fig.3}(a). The starting and destination areas are fixed but the traffic flows vary in different training episodes. The ego vehicle is first spawned randomly within the starting area and follows the planned route to the destination, while avoiding collisions with the surrounding vehicles in the dense traffic. We only consider surrounding vehicles, including motorcycles, sedans, and trucks, as traffic participants and a total of 100 vehicles are randomly spawned in the scene. The surrounding vehicles are running by keeping the target speed (8 m/s) and performing emergency stop when detecting potential danger to the nearby vehicles. The simulation timestep is 0.1 seconds and a training episode ends when encountering the following cases: 1) the ego vehicle reaches the destination area; 2) the ego vehicle collides with other vehicles; 3) the episode exceeds the maximum time steps (800).



We take the bird-eye view image shown in Fig. \ref{fig:fig.3}(b) as the scene representation because it contains rich information of the ego vehicle and its route, the road topology, and surrounding vehicles. We assume that the perception information is perfect, which means the states of the ego vehicle and all surrounding objects, as well as other information on the road, can be projected into a bird-eye view map accurately. The bird-eye view image is an RGB image encoding different information about the driving environment. The drivable areas and lane markings are rendered in grey and white, and the planned route for the ego vehicle is rendered as a thick blue polyline. The historical bounding box trace of the ego vehicle is rendered in red while the historical bounding boxes of the detected surrounding vehicles are rendered as green boxes. The image is with a pixel size of 64$\times$64$\times$3, encoding a field of view with a size of 40$\times$40 $m^2$. The image is aligned to the ego vehicle’s local coordinate where the ego vehicle is positioned at the center.

We decouple the vehicle motion control into longitudinal and lateral directions. Considering that the vehicle basically just needs to follow the drive lane in the lateral direction, to reduce the complexity of the problem and guarantee lateral control stability, the lateral control (steering) is conducted by a pure-pursuit controller to track the target waypoint on the planned route. For the longitudinal control, we utilize two kinds of action spaces for different algorithms. The continuous action space is the normalized throttle and brake control $[-1, 1]$, where $[-1, 0]$ is for brake input and $[0, 1]$ for throttle input, and the discrete action space consists of three actions corresponding to the normalized throttle and brake, which are $\{-1, 0, 1\}$. The policy network employs the convolutional neural network (CNN) structure as the feature extractor and generates the mean and standard deviation of a Gaussian distribution through two fully connected layers, each with 64 hidden units. The critic networks' feature extractors share the same CNN structure. 

In the stage of collecting expert demonstration data, a human expert with a driving license is asked to demonstrate his execution to finish the driving task in the CARLA environment. The expert observes the driving environment from the bird-eye representation with the instant speed information displayed at the same time, and controls the pedal of a Logitech G29 driving set to output continuous actions or presses the keyboard to output discrete maneuvers. It is worth noting that the expertise of the human participant is relative to the RL agent because humans possess prior knowledge on scene understanding and driving with the common goal to drive safely and efficiently. Therefore, the human expert is just asked to drive as usual to finish the task without imposing any other requirements but the speed limit. Overall, a total of 50 trajectories of expert demonstrations are collected for continuous actions, with approximately 15,000 transition tuples. The same amount of trajectories are also collected for discrete maneuvers.

\subsection{Reward function}
Considering the critical factors in urban autonomous driving, we design a reward function that keeps balance on efficiency, ride comfort, and safety. After some trials, the reward function is designed as a combination of four terms:
\begin{equation}
\label{eq16}
r_{t} = r_{v} + r_{step} + r_{col} + r_{safe}.
\end{equation}

The first term $r_{v}$ is for travel efficiency, which stimulates the agent to run as fast as possible but within the speed limit $v_{max}$:
\begin{equation}
\label{eq17}
r_{v} = v + 2 \left( v_{\max} - v \right) \mathds{1} \left(v \ge v_{\max} \right),
\end{equation}
where $v$ is the speed of the ego vehicle. The second term $r_{step}=-0.1$ is a constant step penalty and devised to encourage the agent to complete the task as quickly as possible.

\begin{figure}[htp]
    \centering
    \includegraphics[width=\linewidth]{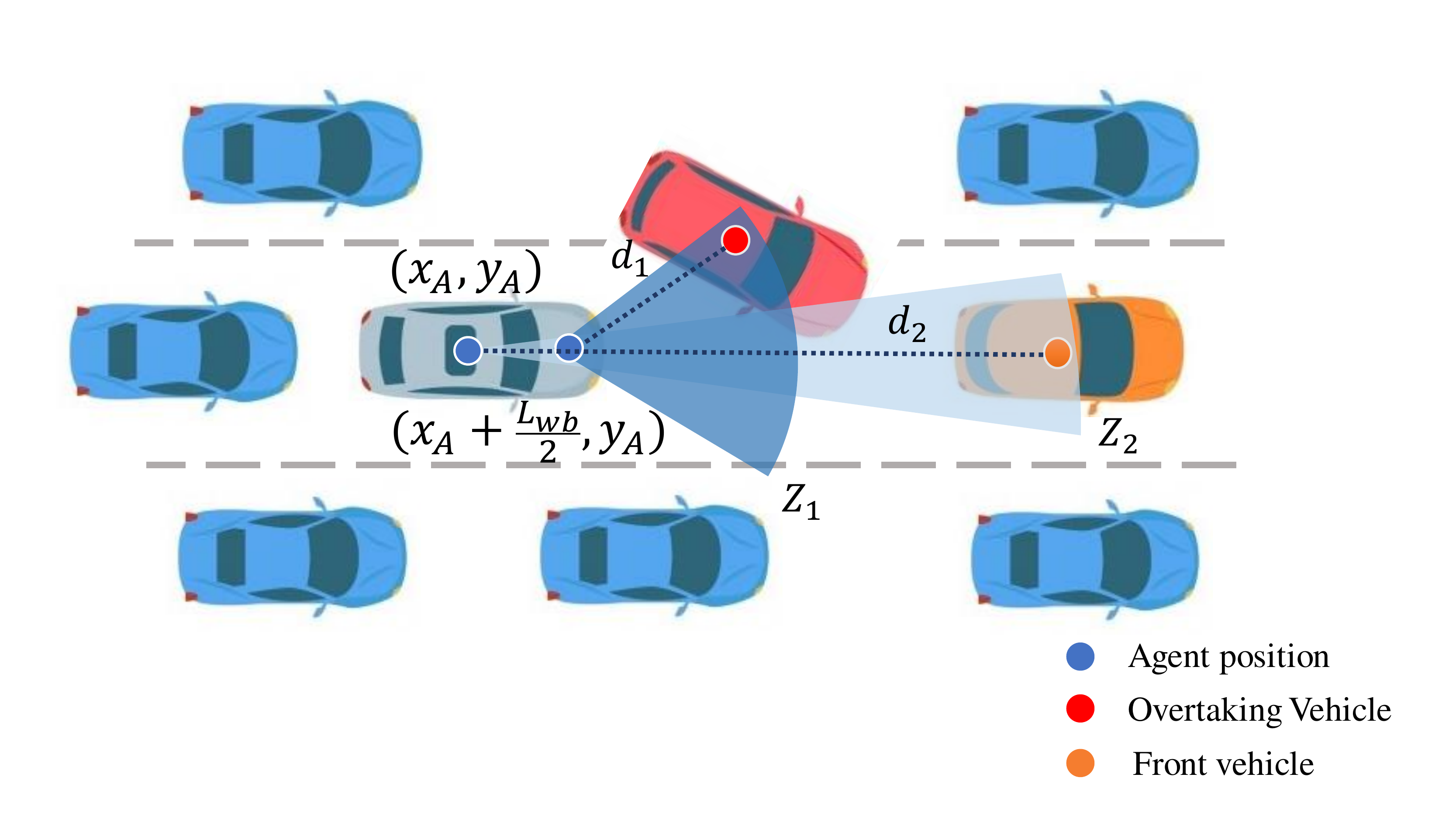}
    \caption{Illustration of the front-detection area for the ego vehicle.}    
    \label{fig:fig.4}
\end{figure}

The other two terms are set for ensuring safety. First of all, $r_{col}=-10$ is a penalty for collision. 
To provide more information to the reward signal, we consider the potential danger in the front-detection zone shown in Fig. \ref{fig:fig.4}. It consists of two fan-shape areas $Z_{1}(\alpha_1, R_1)$ and $Z_{2}(\alpha_2, R_2)$, where $\alpha_i$ and $R_i$ denote its angle and radius. 
These two areas are responsible to provide long- and short-range potential collision information to the reward signal, respectively. If multiple vehicles are in an area $Z_{i}$, only the nearest vehicle is detected and its distance to a certain center point $d_{i}$ will be returned. Thus, the safety reward $r_{safe}$ is given as: 
\begin{equation}
\label{eq18}
r_{\text {safe}} = -\left[ \lambda_{s} \frac{R_1 - d_1}{R_1} \mathds{1} (d_1) + (1 - \lambda_{s} ) \frac{R_2-d_2}{R_2} \mathds{1} (d_2) \right] v_{\text{safe}},
\end{equation}
where $\lambda_{s}=0.8$ balances the importance of the two areas. $v_{\text {safe}}$ is a speed-related regulator. It removes the penalty if the ego vehicle decelerates or waits when facing traffic congestion:
\begin{equation}
\label{eq19}
v_{\text{safe}} = v \left[ 1- \mathds{1} \left(v \le v_{min}, a_{t} < 0 \right) \right],
\end{equation}
where $v_{min}$ is a speed threshold and $a_{t}$ is the agent's action.

\subsection{Comparison baselines}
To make a comprehensive evaluation of the performance of the proposed approach, we compare it with other existing methods. The baseline DRL algorithms are:

1) \textbf{DQN} \cite{21}: a value-based method and Q-function loss gets update through one-step temporal difference error.

2) \textbf{PPO} \cite{40}: an on-policy method and has been widely used in robotics control.

3) \textbf{TD3} \cite{20}: an improved method with twin delayed Q-networks based on the DDPG algorithm.

4) \textbf{A3C} \cite{41}: an on-policy actor-critic framework with asynchronous sampling and advantage estimation. 

5) \textbf{SAC} \cite{22}: the basis of our approach, which has been recently reported with higher performance in practical applications.

The IL baseline methods are listed as follows:

1) \textbf{BC}: a supervised learning method, which has been commonly used in learning driving policy from expert demonstrations.

2) \textbf{SQIL} \cite{31}: a regularized behavioral cloning method, which combines the maximum likelihood of BC with regularization that can stimulate the agent return to demonstrated states upon encountering new states.

3) \textbf{GAIL} \cite{30}: a variant of imitation learning using generative adversarial training, in which the generator is RL-based and the reward function comes from the discriminator that tries to tell apart the demonstrated and generated trajectories.

Moreover, we take \textbf{DQfD} \cite{15} as the reinforcement learning from demonstration baseline method. The DQN, DQfD, and SQIL baseline methods are implemented with discrete action space, and the rest of the baseline methods (PPO, TD3, SAC, A3C, BC, GAIL, and our proposed approach) are implemented with continuous action space.

\subsection{Implementation details}
Our proposed approach and other baseline methods are trained for 100k steps. The neural networks are trained on a single NVIDIA RTX 2070 Super GPU using Tensorflow and Adam optimizer with a learning rate of $3\times10^{-4}$, and the training process takes roughly 6 hours. The parameters related to the experiment are listed in Table \ref{table:tab1}. All the listed algorithms are trained once with the same total training steps and random seeds, and the policy networks are saved after finishing an episode during training. For each algorithm, we take the trained policy network with the highest episodic return for the subsequent testing phase.

\begin{table}[htp]
\caption{Parameters used in the experiment}
\label{table:tab1}
\centering
\begin{tabular}{@{}lllll@{}}
\toprule
Notation                & Meaning                               & Value         \\ \toprule
$v_{max}$               & Speed limit (m/s)                     & 12             \\
$v_{min}$               & Speed threshold (m/s)                 & 0.1           \\
$\alpha_{i=1,2}$        & Angles of front detection area (deg)  & 60, 30        \\
$R_{i=1,2}$             & Radius of front detection area (m)    & 10, 20        \\ 
$L_{wb}$                & Wheelbase (mm)                        & 2850          \\\midrule
$\gamma$                & Discount rate                         & 0.995         \\
$\lambda$               & Polyak averaging weight               & 0.005        \\
$\alpha$                & Initial entropy weight                & 1             \\
$\mathcal{H}_{min}$     & Desired minimum expected entropy      & -1            \\
$N_{\text{buffer}}$     & Agent buffer capacity                 & 50000         \\
$N_{\mathcal{B}}$       & Mini-batch size                       & 64            \\
$\rho$                  & Initial sampling ratio                & 0.3           \\
$\omega$                & Hyperparameter for PER                & 0.6           \\
$\beta$                 & Hyperparameter for PER                & 0.4           \\
$\epsilon$              & Hyperparameter for PER                & $10^{-6}$     \\
\bottomrule
\end{tabular}
\end{table}

\section{Results}
\subsection{Training results}
\begin{figure}[htp]
    \centering
    \includegraphics[width=\linewidth]{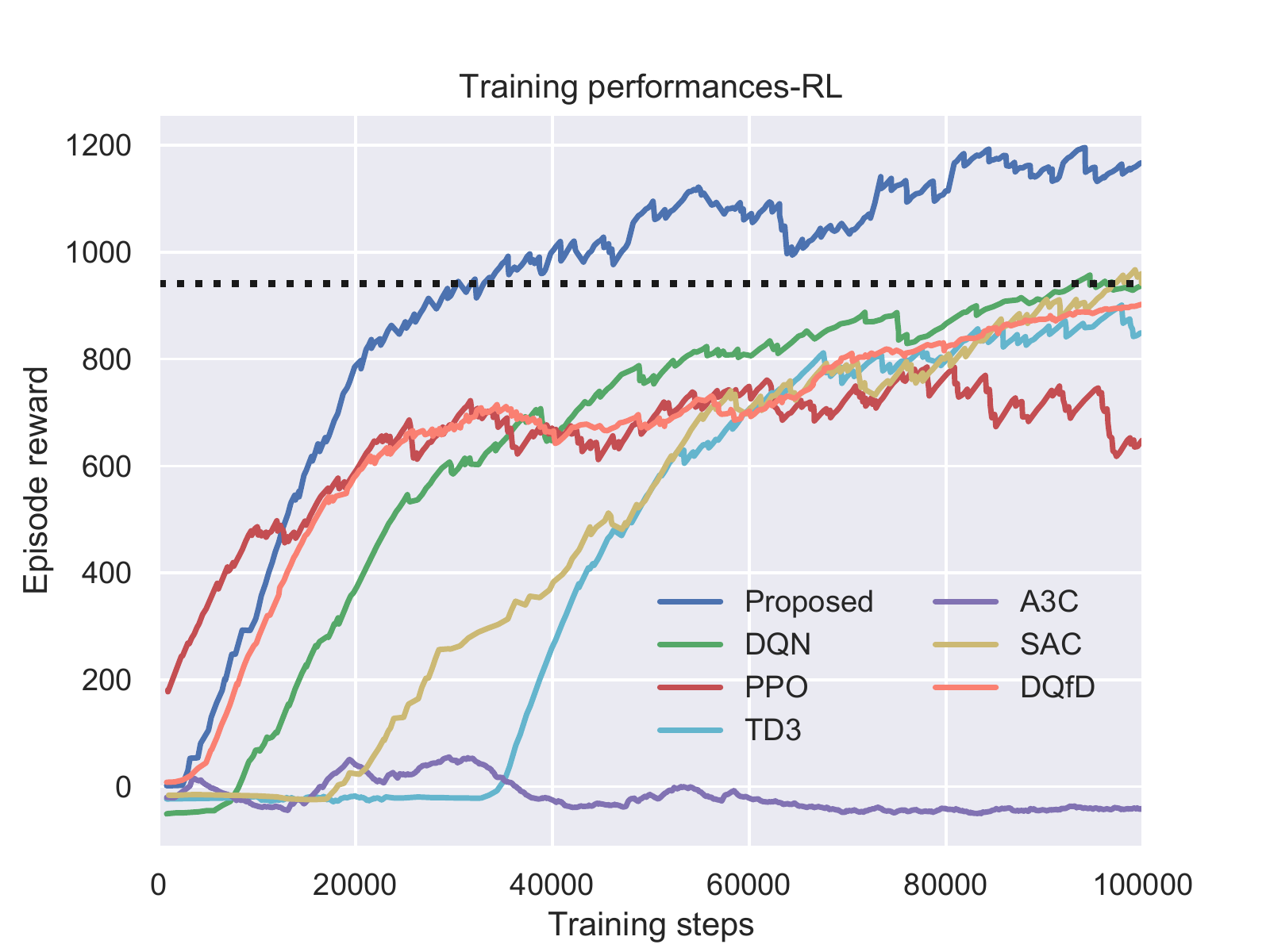}
    \caption{The training curves of RL baselines and our approach. Our approach shows the highest episode reward than other methods at the end of training.}    
    \label{fig:fig.5}
\end{figure}

\begin{figure}[htp]
    \centering
    \includegraphics[width=\linewidth]{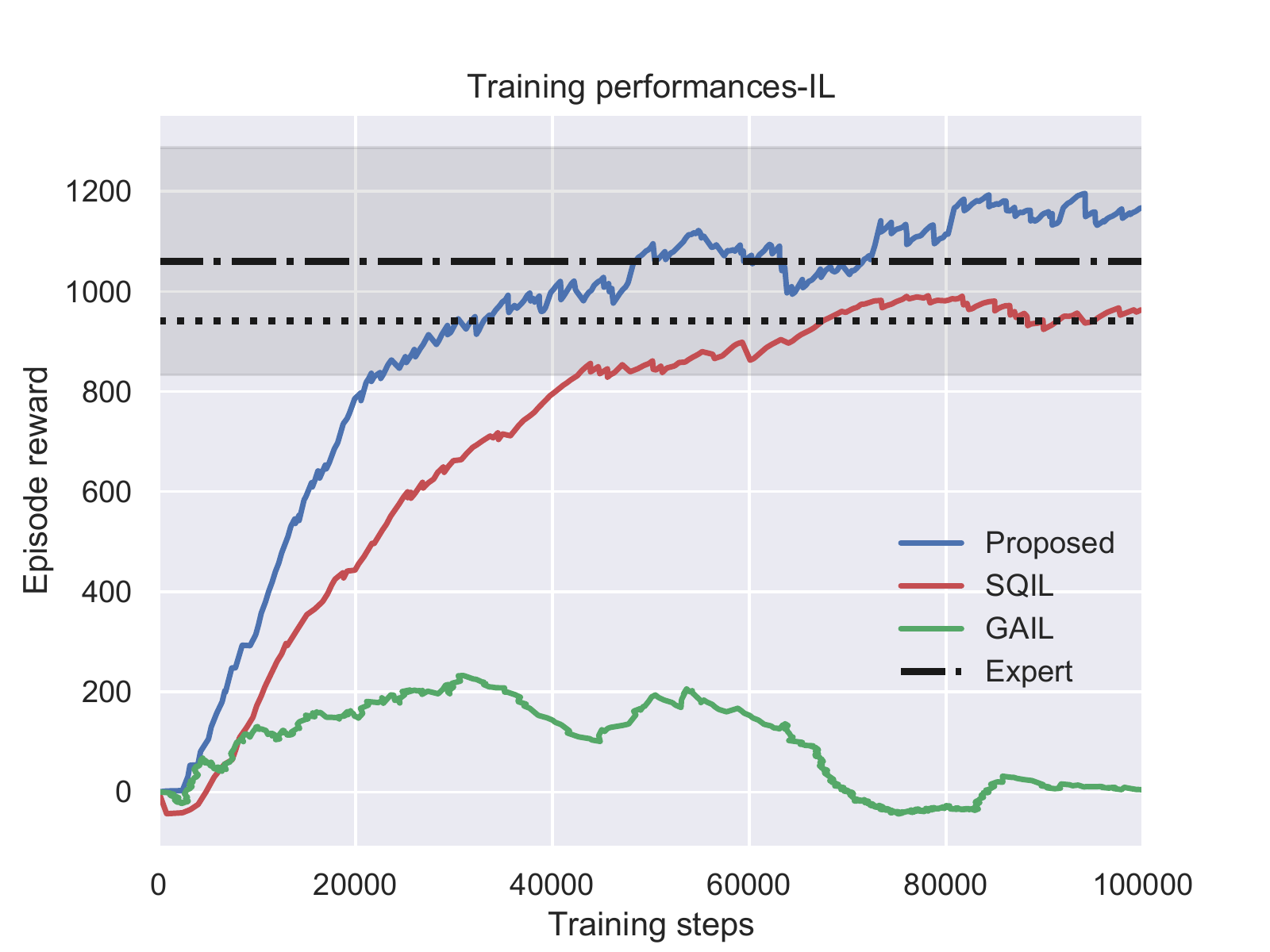}
    \caption{The training curves of IL baselines and our approach. Our approach shows a faster convergence as well as steady progress.}    
    \label{fig:fig.6}
\end{figure}

We evaluate the training performance of our proposed method in comparison with other RL and IL methods introduced in the previous section. Fig. \ref{fig:fig.5} shows the training result of each RL baseline algorithm for 100k training steps, and the black dotted line represents an average episode reward (940.81) that an agent can get reaching the destination area. Fig. \ref{fig:fig.6} shows the training result of the IL algorithms, and the grey shade area in Fig. \ref{fig:fig.6} is the performance of the expert demonstrations with a mean episode reward of 1060.4 (dash-dotted line) and standard deviation of 227.0. 

As seen in Fig. \ref{fig:fig.5}, compared with other RL baseline methods, our proposed method shows a faster convergence speed and better performance at the end. We can find out that the average episodic reward of the proposed method quickly climbs to a very high level and gradually improves and finally converges after roughly 40k steps. At the beginning of the training process, with a larger part of samples from expert demonstration, the agent is basically imitating the expert demonstration with high rewards. Then, with the training going on and the agent's performance getting better, more experience from the agent's self-exploration will be sampled, and thus the training process leans towards reinforcement learning. For the on-policy methods, A3C behaves poorly and can barely reach the entrance of the roundabout. It is because the traffic scenario for each training epoch is notably different, and the A3C policy easily falls into a local minimum of staying close to the starting point. PPO can mitigate this problem to some extent and quickly learn to move forward in the first few episodes, thereby gaining a relatively higher reward at the beginning, but it still cannot learn to reach the destination with dense traffic, showing a limited performance. The three off-policy algorithms perform well and they all can basically reach the destination. SAC performs the best but takes longer steps to converge. DQN actually performs well with only discrete action space due to using double Q-networks, dueling branch, and PER. This is because the value-based method can at least guarantee the training progress, but is heavily relied on the design of the reward function. TD3 takes the longest steps to actually improve the policy, and is very unstable in performance during our experiment: it shows no progress without adding noise, but after adding the noise, its performance still varies greatly. As a closely related method to ours, the DQfD algorithm shows a faster adaptation efficiency and its reward curve converges faster. However, the final performance of DQfD is pretty close to the DQN method and lower than SAC and our proposed method. This is because the policy update of DQfD only depends on the Q values from the Q-network and thus expert demonstrations are not fully exploited, whereas our proposed approach also utilizes expert demonstrations for policy updates. Therefore, the performance of the DQfD baseline cannot reach the same level as our proposed approach.

In comparison with other IL baseline methods, our approach converges quite fast and quickly reaches the level of the human expert. SQIL can also converge very fast but performs notably worse than our method in terms of average episodic reward as it only uses expert demonstration data. GAIL shows the worst performance, probably because it is good at deal with low-dimensional state inputs but fails to handle high-dimensional image inputs, which has also been mentioned by \cite{14} and \cite{31}. The results reveal that our approach can not only imitate the expert demonstrations but also improve the performance to some extent because of adding RL to the framework. For the behavior cloning method, because its learning mechanism is different from other methods (only offline supervised learning), we only show its testing results in the next subsection.

Wrapping up, the performance improvement of our proposed method may come from three factors. First of all, the off-policy RL setting ensures that more abundant and diverse trajectories are stored in the buffer, and the dynamic PER we introduced can make sure that these stored agent trajectories and the expert trajectories can be reused efficiently. Furthermore, we adopt the SAC algorithm that optimizes a stochastic policy through entropy regularization, which brings better exploration capability. Eventually, the introduction of expert demonstration trajectories and imitation learning objectives into the policy training guarantees a lower bound of agent performance, as well as making a faster adaptation for convergence, which is why our proposed method outcompetes other off-policy methods including SAC.

\subsection{Testing results}
The testing scenario is the same as the training scenario in the roundabout but with different traffic conditions. The surrounding traffic participants are randomly spawned in the scene, and each is assigned with a random route in every epoch. We use the trained policy for each method to control the vehicle to navigate through the roundabout.  The success rate, collision rate, average episodic reward, and episodic length are recorded and the summary of test performance of all the methods is given in Table \ref{tab:tab2}. The results reflect that our proposed approach achieves a high success rate, as well as reaching the destination in a shorter time. In line with the training results, A3C does not perform well, and the BC method also suffers from a low success rate due to the distributional shift during the test. SAC and DQN show good performance during the test, and SQIL is also quite effective in solving this problem.

\begin{table}[htbp]
\caption{Summary of the test results}
\centering
\resizebox{\linewidth}{!}{
\begin{tabular}{c|cccc}
\toprule
Method        & Success rate (\%) & Collision rate (\%) & Episode reward               & Episode length (s)          \\ \midrule
BC            & 16                &52                   & 987.37 $\pm$ 248.60          & 48.7 $\pm$ 10.4             \\
A3C           & 18                &49                   & 39.844 $\pm$ 42.764          & 49.3 $\pm$ 2.5             \\
Rule-based    & 60                &40                   & 940.44 $\pm$ 337.21          & \textbf{21.1} $\pm$ 5.0             \\
GAIL          & 65                &26                   & 769.32 $\pm$ 235.47          & 24.5 $\pm$ 6.4             \\
PPO           & 65                &31                   & 903.69 $\pm$ 253.74          & 32.1 $\pm$ 8.3         \\
TD3           & 67                &29                   & 802.45 $\pm$ 184.66          & 31.6 $\pm$ 5.9             \\
SAC           & 76                &14                   & 1050.9 $\pm$ 205.80          & 41.0 $\pm$ 8.4             \\
SQIL          & 78                &12                   & 966.35 $\pm$ 199.13          & 38.2 $\pm$ 8.2             \\
DQN           & 79                &18                   & 1006.3 $\pm$ 191.28          & 38.4 $\pm$ 8.4             \\
DQfD          & 80                &17                   & 1092.6 $\pm$ 149.07          & 35.0 $\pm$ 7.2             \\
Ours          & 90                &8                    & 1205.3 $\pm$ 135.79           & 23.9 $\pm$ 6.5    \\
Ours+Safety   & \textbf{93}       &\textbf{5}           & \textbf{1227.1} $\pm$ 124.05 & 22.8 $\pm$ 5.5    \\
\bottomrule
\end{tabular}}
\label{tab:tab2}
\end{table}

In addition, we add two rule-based methods as comparisons. The first method is a pure rule-based controller that follows the default rule defined by the CARLA simulator. The vehicle will follow the target speed by a PID controller, and will emergently stop if encountering obstacles in a fan shape area like $Z_{i}$ in our reward design. The second one combines our RL-based controller with a rule-based safety controller that will take over the control (emergency brake) if encountering near-collision situations. The results in Table \ref{tab:tab2} indicate that pure ruled-based system does not perform very well. Although the rule-based controller can occasionally finish the task in a shorter time, it suffers from a lower success rate, mainly because of collision with other vehicles. This is because the rules are very simple and the vehicle cannot timely stop to avoid collision with other vehicles. On the other hand, combing RL and rule-based controllers are beneficial in the testing, leading to a higher success rate. When it comes to a real-world scenario, we can design more sophisticated rules to derive a safety control system and combine it with an RL-based control module to ensure both safety and efficiency.

\section{CONCLUSIONS}
In this paper, a novel reinforcement learning algorithm with expert demonstrations is put forward to leverage human prior knowledge, in order to improve the sample efficiency and performance. Specifically, we modify the update of the policy network by combining maximizing the Q-function and imitating the expert's actions, and design an adaptive experience replay method to adaptively sample experience from the agent's self-exploration and expert demonstration for policy update. We validate the proposed method in a simulated challenging urban roundabout scenario with dense traffic. A comprehensive comparison with other RL and IL baselines validates that our method has better sample efficiency and performance in the training process. The testing result reveals that our proposed method can achieve a higher success rate with less time to reach the destination. We also demonstrate that combining a rule-based safety controller with the RL-based controller can further improve the success rate. 
\bibliographystyle{IEEEtran}
\bibliography{b1}
\end{document}